\documentclass[10pt,twocolumn,letterpaper]{article} 
\usepackage{wacv} 
\usepackage{times} 
\usepackage{epsfig} 
\usepackage{graphicx} 
\usepackage{amsmath} 
\usepackage{amssymb} 
\usepackage{booktabs} 

\def\wacvPaperID{1208} 

\wacvalgorithmstrack   

\wacvfinalcopy 

\ifwacvfinal 
\usepackage[breaklinks=true,bookmarks=false]{hyperref} 
\else 
\usepackage[pagebackref=true,breaklinks=true,colorlinks,bookmarks=false]{hyperref} 
\fi

\usepackage{multirow}
\usepackage{multicol}
\usepackage[table]{xcolor}
\usepackage{makecell}
\usepackage[shortlabels]{enumitem}
\usepackage{tabularx}
\usepackage[normalem]{ulem}
\usepackage{float}
\usepackage{graphicx}
\usepackage{enumitem}
\usepackage[accsupp]{axessibility}  

\newcommand\tbf[1]{{\textbf{#1}}}
\newcommand\lv{{\vspace{-0.65em}}}
\newcommand\lvv{{\vspace{-0.5em}}}

\begin{document} 
\title{Transformers For Recognition In Overhead Imagery: A Reality Check}

\author{Francesco Luzi\\ 
Duke University\\ 
{\tt\small francesco.luzi@duke.edu} 
\and
Aneesh Gupta\\ 
Duke University\\ 
{\tt\small aneeshgupta8@gmail.com} 
\and 
Leslie Collins\\ 
Duke University\\ 
{\tt\small leslie.collins@duke.edu} 
\and 
Kyle Bradbury\\ 
Duke University\\ 
{\tt\small kyle.bradbury@duke.edu} 
\and 
Jordan Malof\\ 
University of Montana\\ 
{\tt\small jordan.malof@umontana.edu} 
}

\linespread{1.0}
\maketitle 
\thispagestyle{empty} 

\begin{abstract}  
   
  There is evidence that transformers offer state-of-the-art recognition performance on tasks involving overhead imagery (e.g., satellite imagery). However, it is difficult to make unbiased empirical comparisons between competing deep learning models, making it unclear whether, and to what extent, transformer-based models are beneficial.  In this paper we systematically compare the impact of adding transformer structures into state-of-the-art segmentation models for overhead imagery.  Each model is given a similar budget of free parameters, and their hyperparameters are optimized using Bayesian Optimization with a fixed quantity of data and computation time. We conduct our experiments with a large and diverse dataset comprising two large public benchmarks: Inria and DeepGlobe. We perform additional ablation studies to explore the impact of specific transformer-based modeling choices. Our results suggest that transformers provide consistent, but modest, performance improvements.  We only observe this advantage however in hybrid models that combine convolutional and transformer-based structures, while fully transformer-based models achieve relatively poor performance.      
\end{abstract}

\begin{figure*}[t] 
\begin{center} 
    \fbox{\includegraphics[width=0.90\linewidth]{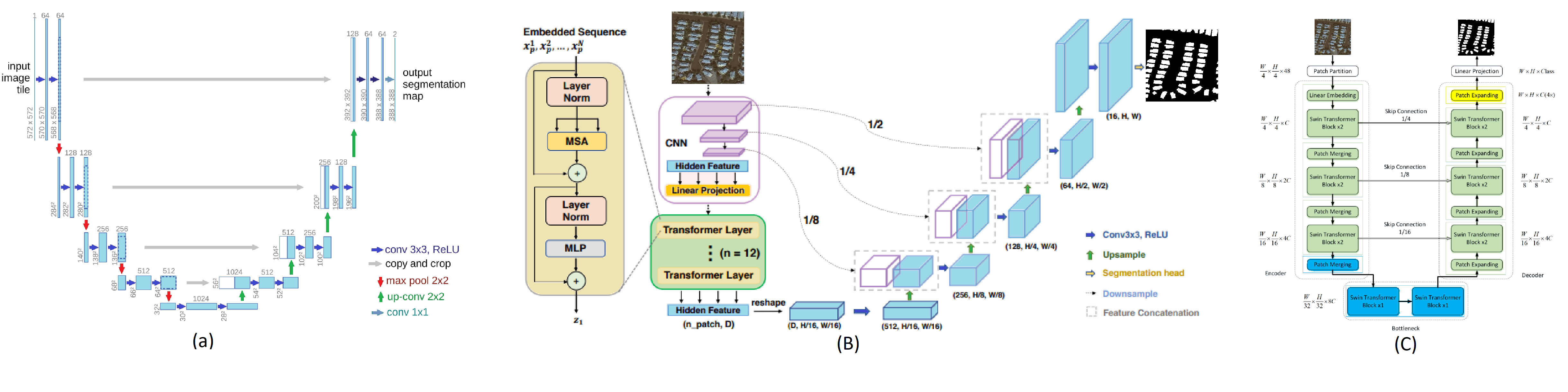} }
\vspace{-0.5em}
\end{center} 
   \caption{(a), (b), and (c) show the architecture for the Unet, TransUnet, and SwinUnet respectively. These figures were inspired directly by \cite{ronneberger2015u, chen2021transunet, cao2021swin}, respectively} 
\label{fig:Arch} 
\vspace{-1.5em}
\end{figure*}

\section{Introduction}
\label{sec:introduction}




\renewcommand{\arraystretch}{0.85}
\begin{table}[tp]
  \begin{center} 
    {
        \begin{tabular}{|c|c|c|c|} 
            \toprule 
            Dataset  & Region & Country  & Size (km$^2$) \\ 
            \midrule 
            \midrule 
            \rowcolor[gray]{0.95}
                  &   Austin& USA         &  81     \\ 
                  &   Chicago& USA        &  81     \\ 
            \rowcolor[gray]{0.95}
            Inria      &   Kitsap County& USA  &  81     \\ 
                  &   West Tyrol& Austria &  81     \\ 
            \rowcolor[gray]{0.95}
                  &   Vienna& Austria     &  81     \\ 
            \hline
              &   Las Vegas& USA      &  150.2     \\ 
            \rowcolor[gray]{0.95}
            \multirow{2}{*}{DeepGlobe}   &   Paris& France       &  41.88     \\ 
              &   Shanghai& China     &  173.32     \\ 
            \rowcolor[gray]{0.95}
              &   Khartoum& Sudan     &  32.88     \\ 
            \bottomrule 
        \end{tabular} 
    }
\end{center} 
\vspace{-0.5em}
\caption{The cities and their size that compose the Inria and DeepGlobe datasets.} 
\label{DataTable}
\vspace{-1.5em}
\end{table}

Transformer-based models have become prevalent in computer vision tasks and have achieved state-of-the-art performance in classification \cite{dosovitskiy2020image, liu2021swin}, object detection \cite{dai2021dynamic, zhang2022dino}, and segmentation \cite{chen2021transunet, cao2021swin}.  This success might ostensibly suggest that transformers are superior to other existing models, such as those based upon convolutional structures, however this is difficult to conclude based upon the existing research literature due to the absence of experimental controls when comparing different vision models.  The performance of modern vision models - all of which are based upon deep neural networks - are affected by numerous factors that vary widely among competing models used in public benchmarks, and in the research literature \cite{musgrave2020metric}.  This includes factors such as the quantity and quality of training data, the training algorithm (e.g., optimizer), the training time allotted, and the model's size (i.e., the number of free model parameters).  Another more subtle, but highly influential factor, is the computation time and effort invested by the designer on hyperparameter optimization, which can result in misleading performance comparisons \cite{musgrave2020metric}.  

If one or more of the aforementioned factors vary between competing vision models, then it is unclear which factors among them are responsible for any performance differences \cite{renard2020variability, islam2017reproducibility, hutson2018artificial, musgrave2020metric}. Consequently, it is unclear whether the recent success of transformer-based models applied to overhead imagery has been driven by the use of transformers, or the variety of other factors that vary among the competing models.   A major goal of vision research is to uncover the underlying causal factors and design principles that underpin vision systems; this not only advances our understanding of vision systems, but also often leads to substantive performance improvements in such systems.  Therefore an important question in the vision literature is whether, and to what extent, transformers generally benefit vision models. Controlled studies of transformers have been conducted with natural imagery \cite{paul2022vision}, providing some evidence in that context.  However, it is unclear whether their success extends to the unique statistics and conditions present in overhead imagery, a major area of vision research. Transformers excel at modeling long range dependencies, which while generally beneficial in most vision tasks, may not be as important in segmentation of overhead imagery where building information is compact, highly localized, and many times isolated from other structures.  To our knowledge there has been no systematic study of this question for overhead imagery tasks. 

In this work we perform a carefully-controlled empirical comparison of three state-of-the-art segmentation models using overhead imagery, where each model utilizes progressively more transformer-based structures.  Specifically, we consider the following three models: U-Net \cite{Iglovikov_2018_CVPR_Workshops}, TransUNet\cite{chen2021transunet}, and SwinUnet\cite{cao2021swin}. This is the first time the TransUnet and SwinUnet have been applied to a large-scale dataset of overhead imagery \footnote{The recent work in \cite{wang2022unetformer} independently and concurrently studied these models, in a complementary setting}. Aside from the model variation we carefully control all other experimental factors, such as the size of the models, their quantity of training data, and training procedures.  We use a large and diverse dataset of overhead imagery, comprising two publicly-available benchmarks to maximize the generality and relevance of our results. To provide a transparent and unbiased hyperparameter optimization procedure, we use Bayesian Optimization (BO) with a fixed budget of iterations to select the hyperparameters of each model.  We provide each model with approximately 330 hours of optimization time in order to identify effective hyperparameters for each model. These experimental controls allow us to study whether, and to what degree, transformers are beneficial in the context of overhead imagery.  Using our optimized models, we also conduct several additional ablation studies to evaluate the impact of specific design choices in the transformer-based models. We can summarize our contributions as follows: 
\begin{itemize}
    \item\lv The first investigation of two recent state-of-the-art segmentation models for processing overhead imagery: the transUnet \cite{chen2021transunet}, and the swinUnet \cite{cao2021swin}.    
    \item\lv The first controlled evaluation of whether, and to what extent, transformers are beneficial for vision models in overhead imagery.  \lv
\end{itemize}

\section{Related Work}
\label{sec:related_work}
\tbf{Segmentation in overhead imagery.}
Segmentation of overhead imagery requires complex features to describe the vast domain as well as pixel level precision. Initially developed for medical imagery, Unet \cite{ronneberger2015u} has been show to be a powerful model in overhead image segmentation \cite{huang2018large, he2020hybrid} and in the broader segmentation community, with many variations on the model such as Dense-Unet \cite{li2018h}, Res-Unet \cite{xiao2018weighted}, Unet++ \cite{zhou2018Unet++}, V-Net\cite{milletari2016v}, and Unet3+ \cite{huang2020Unet}. This is attributed to the auto encoder-like structure where it receives its "U" shape and name combined with the skip connections, feeding high resolution spatial information into the last layers of the model.

Other models such as DeepLabv3 \cite{chen2018encoder} and Mask-RCNN \cite{he2017mask} have been used successfully in segmentation of overhead imagery \cite{kong2020synthinel, carvalho2020instance}. While these models also perform very well, we chose to evaluate Unet-based architectures due to the high performance and large number of variant models. This high number of variants allowed us to more easily compare small changes in the model architecture. 

\tbf{Transformers in segmentation.}
Very recently transformers have started to be used in that segmentation of overhead imagery \cite{he2022swin, sun2022multi}, achieving good performance. Transformers had already started to become common in other domains such as TransUnet \cite{chen2021transunet}, ViT-V-Net \cite{chen2021vit}, TransClaw U-Net \cite{chang2021transclaw}, UTNet \cite{gao2021utnet}, Cotr \cite{xie2021cotr}, and SwinUnet \cite{cao2021swin} in medical image segmentation. 

\tbf{Evaluation of transformers.}
Transformers are rather new in computer vision and have only recently become state-of-the-art. As a result, their impact on performance has not been thoroughly analyzed in many domains and applications, including our own. While work has been done in evaluating their generalization capabilities in respect to distribution shift \cite{zhang2022delving} and how transferable their learned representations are \cite{zhou2021convnets}, these are very general results about the feature representations derived for other applications. For segmentation of overhead imagery, to our knowledge, there has been no work done towards thoroughly evaluating and isolating the effect of transformers in state-of-the-art models.



\begin{figure*}[tp] 
\begin{center} 
\fbox{ \includegraphics[width=0.27\textwidth]{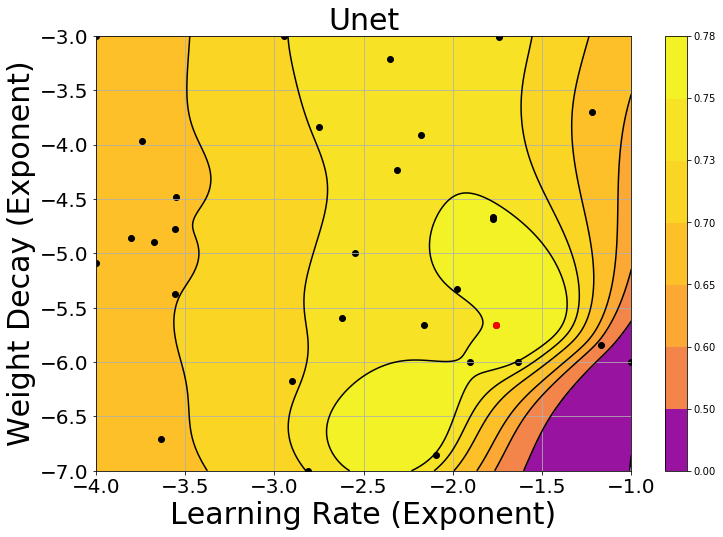} }
\fbox{ \includegraphics[width=0.27\textwidth]{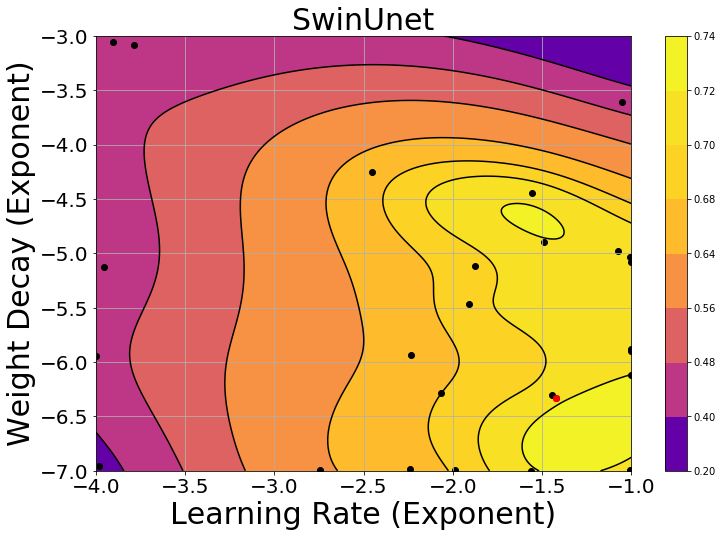} }
\fbox{ \includegraphics[width=0.27\textwidth]{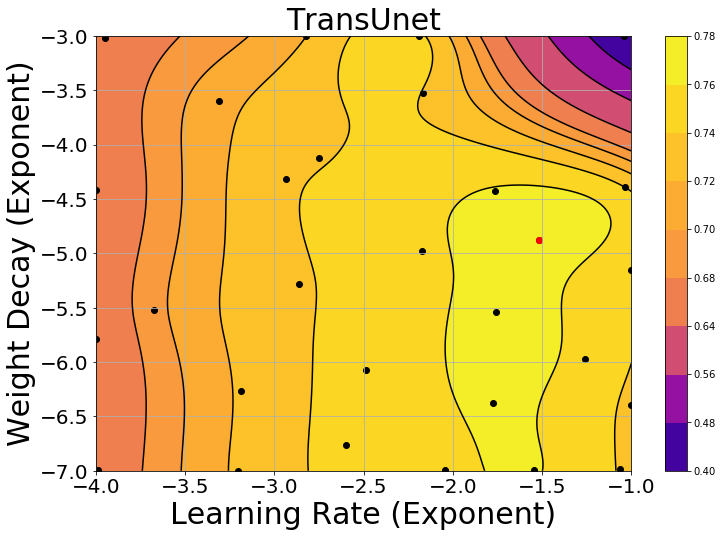} }
\vspace{-0.5em}
\end{center} 
   \caption{Heat maps of the parameter space searched with Bayesian Optimization. Sampled points are displayed as block dots, with the final parameter choice represented by a red dot, and Gaussian processes are used to model the points in between to fill out the space. Learning rate is plotted on the X-axis with weight decay plotted on the Y-axis. Both learning rate and weight decay are sampled by their exponential ($-3$ gives a value of $10^{-3}$ or $1e-3$) so that there is equal weight given to parameter values on a $\log_{10}$ scale.} 
\label{fig:BO_Heatmap} 
\vspace{-1.5em}
\end{figure*}

\lvv
\section{Benchmark datasets}
We train and evaluate our data on two large publicly-available datasets of overhead imagery: the DeepGlobe (DG) Competition Dataset \cite{demir2018deepglobe}, and the Inria Building Labeling Competition Dataset \cite{maggiori2017can,huang2018large}. Both of these datasets contain high-resolution (0.3m ground sampling density) color imagery with pixel-wise labels indicating the presence of a building (a value of one), or not (a value of zero).  Collectively, these datasets encompass nine diverse cities spanning North America, Europe, and Asia, as summarized in Table \ref{DataTable}.


\lvv
\section{Benchmark Segmentation Models}
Our aim is to compare similar state-of-the-art models, shown in Fig \ref{fig:Arch}, and determine what factors contribute to the overall performance. Specifically, we aim to answer whether or not transformer layers are valuable in overhead segmentation. For this we selected the Unet \cite{ronneberger2015u}, TransUnet \cite{chen2021transunet}, and SwinUnet \cite{cao2021swin} models since they all use the same base Unet structure, they are all state-of-of-the-art models in segmentation, and they contain different levels of transformer integration. SwinUnet being entirely transformer based, Unet being completely convolutional, and TransUnet a hybrid of transformer and convolutional neural networks.

For most of our comparisons we restrict the number of parameters for each model such that they can be compared fairly. The Unet and TransUnet natively have a parameter count around 105 million and so we select this for our baseline models. A brief description of each model and their importance is given below, a more detailed description can be found in the supplemental material.

\begin{figure}[tp] 
\begin{center} 
\fbox{ \includegraphics[width=0.45\textwidth]{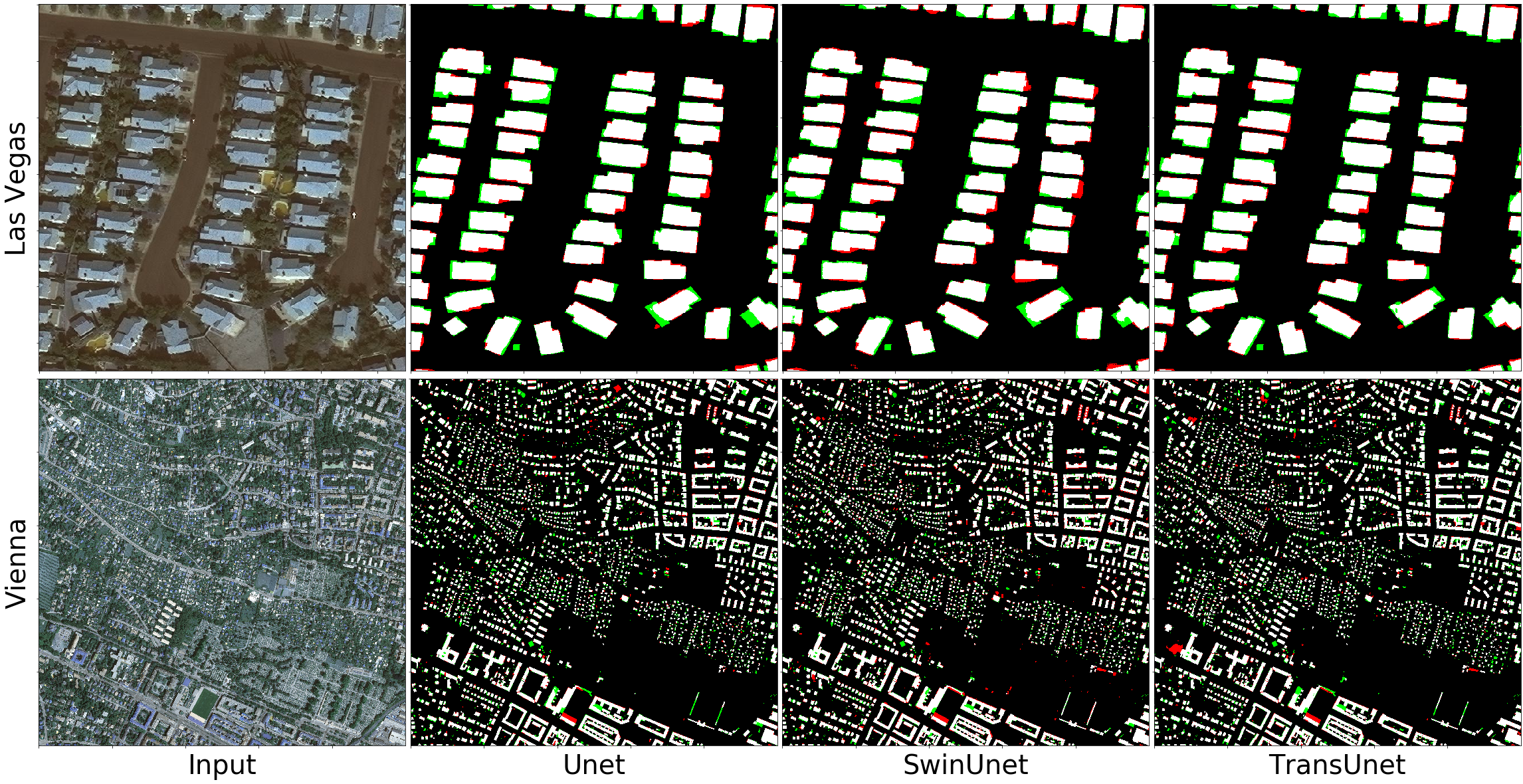} }
\vspace{-1.5em}
\end{center} 
   \caption{ Examples of input images and mask outputs. Each row is a test image from a different city, with the first row from DG and the second from Inria. The first column contains the input image, the next three columns contain the predictions from the Unet, SwinUnet, and TransUnet respectively. The difference with the ground truth is highlighted, with green showing missed building pixels and red denoting false alarm building pixels.} 
\label{fig:Example} 
\vspace{-1.5em}
\end{figure}

\textbf{Unet.} For our baseline Convolutional model we used a Unet based architecture with a ResNet101 \cite{he2016deep} backbone, shown in Fig \ref{fig:Arch} (a). The model uses ResNet101 weights pretrained on ImageNet \cite{krizhevsky2012imagenet} with the standard Unet structure described in the paper. We chose to use the Unet due to its popularity and performance in segmentation tasks \cite{ye2021ct, adiba2019transfer, huang2018large, iglovikov2018ternausnet}. Its simple and intuitive design has allowed for many variants, including the transformer based variants we considered. 

\textbf{TransUnet.} For the majority of our experiments we use a standard TransUnet with Visual Transformer (ViT) blocks \cite{dosovitskiy2020image} pretrained on ImageNet, shown in Fig \ref{fig:Arch} (b). The official implementation is used for the model, and all model changes are derived from that code base. The TransUnet proved to be an excellent choice for a convolution-transformer mixed model due to its simple modification. The TransUnet is very similar to the standard Unet but with transformer layers in the deepest part of the encoder. This allows for an ablation over different aspects of the model without effect on the rest of the layers. 

\textbf{SwinUnet.} Our fully transformer-based model is a modified SwinUnet, shown in Fig \ref{fig:Arch} (c), with weights loaded from a Swin Base model \cite{liu2021swin}. We used the implementation from the original code base for all of our work. We had modified the original SwinUnet architecture to achieve an equivalent parameter count to the other models. We also train a SwinUnet with a Swin Tiny backbone architecture as is described in the SwinUnet paper and compare that to a Unet with equivalent parameters to verify that our modified SwinUnet is representative of the model's performance. We use the SwinUnet to represent fully transformer based models due to its state-of-the-art performance in medical segmentation and the Swin transformers dominance in other computer vision based tasks such as classification and object detection \cite{liu2021swin, dai2021dynamic}. 

\lvv
\section{Experimental Design}
\label{sec:benchmark_design}
The primary goal of our study is to compare the effectiveness of recent transformer-based models to state-of-the-art convolutional models, while controlling the number of trainable model parameters.  

\subsection{Data Handling}
While the DG dataset contained labels for multiple classes (e.g., road, building, etc.), we only used the building labels so that we can train on both datasets together. Our combined dataset involves a two thirds, one sixth, one sixth split for training, validation, and testing respectively. For Inria, we use the official test set as our test set (first six tiles) and for DG we randomly select one sixth of the data for the test set. Each city has the same proportions represented in the training, validation, and test sets.

\setlength{\tabcolsep}{6pt}
\begin{table}[tp]
  \begin{center} 
    {\small{ 
        \begin{tabularx}{\columnwidth}{|X|X|} 
            \toprule 
            Parameter  & Range \\
            \midrule 
            \midrule 
            \rowcolor[gray]{0.95}
            Learning Rate       & $10^\mu$, $\mu \in U(-4, -1)$ \\ 
            Weight Decay        & $10^\mu$, $\mu \in U(-7, -3)$  \\ 
            \rowcolor[gray]{0.95}
            Window Size         & [2, 4, 8] \\ 
            \bottomrule 
        \end{tabularx} 
    }} 
\end{center} 
\vspace{-0.5em}
\caption{ Here we list the parameters used in BO in the first column and their possible values in the second column. $\mu$ is taken from a uniform distribution. Learning rate and weight decay are found using Bayesian optimization in all models and window size is only used in the Bayesian optimization for the SwinUnet. } 
\label{tbl:HypUsed}
\vspace{-1.5em}
\end{table} 
\setlength{\tabcolsep}{6pt}


\subsection{Model hyperparameter optimization}
\label{subsec:model_optimization}
To minimize bias towards a particular model, we optimized all competing models using BO \cite{snoek2012practical}, which is a systematic, replicable, and transparent process to search for optimal hyperparameters via experimentation. Furthermore, each model was allowed $30$ iterations of BO, ensuring comparable computational resources were provided to each model. We chose to use $30$ iterations based on our experiments with the Unet. We found that $30$ trials explored the hyperparameter space thoroughly and any more trials would give little to no performance improvements. After approximately the 20th iteration of BO, the models converged to a local minima and had minor to no improvements with continued search time. Fig \ref{fig:BO_Heatmap} displays a predicted heat map of the parameter search space along with the points sampled during BO. In the supplemental work we include figures that show the model performance converging, training validation for each BO iteration, and the model performance corresponding to individual parameters selected. We also had limited computational resources and $30$ trials took approximately two weeks to complete, requiring six weeks in total to optimize over the three models. For BO we used the python implementation provided by the BO library \cite{BaysOpt}. We initialized the BO with $2$ random points, used the expected improvement \cite{jones1998efficient} acquisition function and set the exploitation-exploration trade-off parameter ("xi") to 0.1, which we found to work well through initial trial and error. All other parameters were remained at their default values.

For each of our three model classes, we identified a small set ($2$-$3$) of the most influential hyperparameters for inclusion in the BO, shown in Table \ref{tbl:HypUsed}. This includes learning rate and weight decay, with the addition of window size for the SwinUnet models. Our goal was to select the most impactful hyperparameters that would not change the number of parameters available to the model.  In each case we trained and tested with randomly-selected hyperparameter settings to initialize the Gaussian Process in the BO. The models were trained using one fifth of the training dataset to expedite the process. Note that this still resulted in a relatively large training dataset, including $~107$ km$^2$ of satellite imagery spanning $9$ regions, and requiring 12 hours per model trained.  The last epoch performance on the validation set was used as the target parameter to optimize in the BO. The hyperparameters with the highest validation performance were then adopted to train a model on the full training dataset, which was then evaluated on the withheld test set as a final unbiased estimator of the model's performance. 

The transformer-based models have additional impactful hyperparameters, such as the number of heads, embedding dimension, and layer count, however as discussed above, these can alter the number of trainable parameters in the model and therefore we fixed these parameters in advance.  We kept these parameters consistent with what was used in the pretrained models whose weights we use to initialize our models. To avoid disadvantaging any model, we did not include these parameters in the BO. However, after performing BO, we study the impact of the number of transformer layers in a model by searching over a grid of settings. We do not explore these for the other architectural parameters since it would remove the ability to use pretrained weights.

\setlength{\tabcolsep}{6pt}
\begin{table} 
  \begin{center} 
    {\small{ 
        \begin{tabular}{|l|c|c|c|c|} 
            \toprule 
            \multirow{2}{*}{Model}   & \multirow{2}{*}{Inria}  & \multirow{2}{*}{DG} & \multirow{2}{*}{Composite}  & Parameters \\ 
                                     &   &      &     & (M) \\ 
            \midrule 
            \midrule 
            \rowcolor[gray]{0.95}
            Unet34          & \tbf{76.58} & \tbf{78.71} & \tbf{77.74} & 26.71 \\ 
            SwinUnet Tiny        & 75.67 & 77.85 & 76.87 & 27.13 \\ 
            \bottomrule 
        \end{tabular} 
    }} 
\end{center} 
\vspace{-0.5em}
\caption{Smaller versions of Unet and SwinUnet were also trained to verify that the performance gap exists with the original SwinUnet architecture} 
\label{SmallResults}
\vspace{-1.5em}
\end{table} 
\setlength{\tabcolsep}{6pt}

\setlength{\tabcolsep}{2pt}
\begin{table*} 
 \begin{center} 
    {\small{ 
        \begin{tabular}{|l||c|c|c|c|c||c|c|c|c||c||c|} 
            \toprule 
            \multirow{2}{*}{Model}  &  \multicolumn{5}{|c|}{Inria} &     \multicolumn{4}{|c|}{DeepGlobe}  & \multirow{2}{*}{Composite}  & Parameters \\ 
                                             & Austin  & Chicago & Kitsap County & West Tyrol & Vienna  &   Las Vegas & Paris & Shanghai & Khartoum &    & (M) \\ 
            \midrule 
            \midrule 
            \rowcolor[gray]{0.95}
            Unet \cite{ronneberger2015u}       & 81.92  & 71.58   &  69.00        & 80.44      & 82.53  &  \tbf{85.50}      & 72.26  &  77.13   & 73.64    &  79.53   & 104.89 \\ 
            SwinUnet \cite{cao2021swin}        & 80.21  & 69.62   &  68.70        & 80.33      & 81.85  &  84.87      & 70.57  &  76.25   & 72.36    &  78.48   & 102.64 \\ 
            \rowcolor[gray]{0.95}
            TransUnet \cite{chen2021transunet} & \tbf{81.94}  & \tbf{73.21}   &  \tbf{69.19}        & \tbf{81.46}      & \tbf{82.94}  &  85.45      & \tbf{72.92}  &  \tbf{77.31}   & \tbf{73.79}    &  \tbf{79.96}   & 105.91 \\ 
            \bottomrule 
        \end{tabular} 
    }} 
\end{center} 
\vspace{-0.5em}
\caption{Comparison of performance of the three model architectures we tested broken down by city, measured by the intersection-over-union (IoU). The Unet represents the convolutional only approach, SwinUnet the transformer only approach, and TransUnet being a combination of the two. Model architecture was selected so that the models would have roughly similar parameter counts. The best performance on each city is shown in bold.} 
\label{tbl:BaseResults}
\vspace{-1.5em}
\end{table*} 
\setlength{\tabcolsep}{6pt}

\subsection{Training}
\label{subsec:training}
All models were trained using the hyperparameters found from BO on the base model unless stated otherwise. For example the TransUnet with $6$ transformer layers used the same parameters as were found using the TransUnet with $12$ layers. This approach was used to decrease search time, and we observed that hyperparameters shared between similar architectures perform very well. Since training on a subset of the data was utilized to save time, we used an adaptive learning schedule and training time. This avoided the issue of selecting a learning schedule that may have unfairly advantaged one model over another, as well as making the transition from utilizing a subset of the training data set to utilizing the full data set smoother. We allowed the models to train until the validation curve flattened and then reduced the learning rate by a factor of $2$. The learning rate was dropped three times and then the training was halted. To determine if the validation had flattened we maintained a running average of the validation performance and compared the current value against the running average obtained $10$ epochs prior. Once there was zero difference between the two, or the difference was negative, the learning rate was lowered. After the third drop in learning rate, training was continued until the validation curve flattened once more and then training was halted with the last epoch validation used for scoring the parameter selection.

All models were trained, validated, and tested on Inria and DG. $650 \times 650$ patches were taken from each satellite image, normalized by the dataset statistics, and cropped randomly during training to be $512 \times 512$. Random rotations of $90^\circ$ were also used during training. The models were evaluated on a pixel-wise cross entropy loss and a soft intersection over union (IoU) \cite{rezatofighi2019generalized} loss, weighted equally. \lvv

\subsection{Performance Metrics}
We report performance using intersection-over-union (IoU), because it is widely-used for segmentation of overhead imagery, and is the official performance metric for the Inria and DG datasets \cite{demir2018deepglobe, maggiori2017can}.  IoU measures the intersection of all predicted building pixels and ground truth labels over the union of all predicted building pixels and ground truth labels. The IoU is given by,
\begin{equation}
    IoU = \frac{Prediction \cap Labels}{Prediction \cup labels}
\end{equation}
This results is a metric normalized between $0$ and $1$ that intuitively encapsulates how well the model predicted the ground truth labels. The three right columns of images in Fig \ref{fig:Example} demonstrate an example of prediction vs. ground truth where the intersection is shown by the white pixels and the union is all non-black pixels.

\lvv
\section{Model Performance Comparisons}
\label{sec:results_performance_comparisons}
We performed a number of experiments to isolate specific changes and impact of the model and parameter selection process. We evaluated the effect that different numbers of transformer layers, pretraining, parameter searches, and transformer layers in general had on the model's performance. Fig \ref{fig:Example} demonstrates each model's performance on our test sets.

\textbf{Standard-sized models.} The results from our standard-sized models are reported in Table \ref{tbl:BaseResults}, where  TransUnet achieves the highest overall IoU, followed by the Unet and the SwinUnet, respectively.  These rankings persist for each of the two benchmark datasets as well, suggesting that the results are somewhat robust to variations in the underlying data. These results are comparable to current state-of-the-art results on Inria \cite{chen2021building, kang2021picoco, zorzi2021machine}.  The results \textit{tentatively} suggest that including some transformer modules (e.g., the TransUnet) is beneficial, however including too much can be detrimental  (e.g., SwinUnet).  The TransUnet performs consistently better than the Unet, which reflects the finding in the TransUnet and SwinUnet papers \cite{chen2021transunet, cao2021swin}, but the SwinUnet results seem to under-perform. 
Prior research has found that Transformer-based models are more difficult to train than convolutional models, and tend to improve faster with growing quantities of training data \cite{liu2021efficient, dosovitskiy2020image}.  It is possible therefore that, as overhead imagery datasets continue to grow, the SwinUnet may perform relatively better.  Given the sample complexity and size of our training data however, it does not.  The lower performance of SwinUnet in this setting provides support to the notion that long range dependencies are not as beneficial in the early stages of segmentation in overhead imagery and that localized information is important in overhead imagery. 

\textbf{Small-sized models.} To fairly compare the SwinUnet model and increase its parameter count we added transformer blocks and loaded the model from the pretrained Swin Base checkpoint, using the base architecture instead (e.g. number of heads, hidden layer dimensions). This could disadvantage the model since we have not performed an extensive architectural search for the optimal large SwinUnet, whereas, the Unet and TransUnet were tested in their original form and thus were likely more optimized. To verify that in enlarging the SwinUnet architecture we didn't disadvantage the model in training we also compared the SwinUnet with a Swin Tiny backbone to the Unet with ResNet34 structure. We performed BO on both these models in the same manner as described above and trained them in exactly same manner. Table \ref{SmallResults} shows the results on the test set and the results are consistent with the results found using the larger models as reported in Table \ref{tbl:BaseResults}. 

\subsection{Are Transformers Beneficial?}


Under fair conditions TransUnet, a transformer-convolutional hybrid, out-performed both the alternate Unet and SwinUnet on a large and diverse test set. While this gives credibility to the notion that transformers improve a model's performance for segmentation of overhead imagery, this claim cannot be made based on the performance improvements alone. We have found that while removing all transformer layers from TransUnet decreases performance, as shown in Table \ref{LayerResults}, the model still outperforms the small Unet and performs similarly to the large Unet. Since the TransUnet with no transformer layers and a greatly reduced model size (by parameter count) performs on a similar level to the Unet then it is clear that there are other factors in the TransUnet architecture that boost performance. To resolve this ambiguity, we performed an ablation by substituting the transformer layers in the TransUnet with other layer types, shown in Table \ref{TranResults}.
We found that while the benefit is small, adding transformers can improve model performance over other layer types.


\lvv
\section{Additional Analysis and Ablations}
\label{sec:results_additional_analysis}

\subsection{Window Size Effect}
One of the parameters used for the SwinUnet in the BO search was the window size. We found in the search that a window size of $4$ was ideal for performance but wanted to verify the accuracy of the search and the importance of this parameter. In Table \ref{WinResults} we provide the performance of $3$ models trained with window size of $2$, $4$, and $8$. The results show that the selected window size performed best and that increasing the window size had negligible effects while reducing it caused a large drop in performance.

\setlength{\tabcolsep}{4pt}
\begin{table}
  \begin{center} 
    {\small{ 
        \begin{tabular}{|l|c|c|c|c|c|} 
            \toprule 
            \multirow{2}{*}{Model} & Window  & \multirow{2}{*}{Inria}  & \multirow{2}{*}{DG} & \multirow{2}{*}{Composite}  & Parameters \\ 
                                   &  Size   &   &      &     & (M) \\ 
            \midrule 
            \midrule 
            \rowcolor[gray]{0.95}
               &    2     & 76.90 & 78.95 &  78.02 & 102.62 \\
            SwinUnet   &    4     & \tbf{77.18} & \tbf{79.55} & \tbf{78.48} & 102.64 \\
            \rowcolor[gray]{0.95}
               &    8     & 77.12 & 79.10 &  78.21 & 102.73 \\ 
            \bottomrule 
        \end{tabular} 
    }} 
\end{center} 
\vspace{-0.5em}
\caption{We trained the SwinUnet with varying window sizes to determine the effect of window size on the model and to verify that the BO correctly picked the highest performing parameter. Each row provides the performance on the test set of a SwinUnet model trained with a different window size. All other training parameters were kept constant. } 
\label{WinResults}
\vspace{-0.5em}
\end{table} 
\setlength{\tabcolsep}{6pt}

\setlength{\tabcolsep}{2.5pt}
\begin{table}[tp]
  \begin{center} 
    {\small{ 
        \begin{tabular}{|l|c|c|c|c|c|} 
            \toprule 
            \multirow{2}{*}{Model} & Transformer  & \multirow{2}{*}{Inria}  & \multirow{2}{*}{DG} & \multirow{2}{*}{Composite}  & Parameters \\ 
                                   &  Layers      &   &      &     & (M) \\ 
            \midrule 
            \midrule 
            \rowcolor[gray]{0.95}
            Unet34     &    0     & 76.58 & 78.71 &  77.74 & 26.71 \\
            Unet101    &    0     & 78.41 & 80.46 &  79.53 & 104.89 \\ 
            \hline
            \rowcolor[gray]{0.95}
              &    0     & 78.65 & 80.20 &  79.51 & 20.86 \\
              &    6     & 80.02 & 80.83 &  80.47 & 63.38 \\ 
            \rowcolor[gray]{0.95}
            \multirow{2}{*}{TransUnet}  &    8     & \tbf{80.12} & 81.04 &  \tbf{80.63} & 77.56 \\ 
              &    10    & 79.90 & 80.94 &  80.47 & 91.74 \\ 
            \rowcolor[gray]{0.95}
              &    12    & 79.22 & 80.56 &  79.96 & 105.91 \\ 
              &    14    & 79.98 & \tbf{81.07} &  80.58 & 120.09  \\ 
            \bottomrule 
        \end{tabular} 
    }} 
\end{center} 
\vspace{-0.5em}
\caption{We performed an ablation over the number of transformer layers used. Unet and TransUnet with 0 layers are included to show the benefit that transformer layers provide.} 
\label{LayerResults}
\vspace{-1.5em}
\end{table} 
\setlength{\tabcolsep}{6pt}

\setlength{\tabcolsep}{10pt}
\begin{table*}[tp]
\begin{center} 
    {\small{ 
        \begin{tabular}{|l|c|c|c|c|c|c|} 
            \toprule 
            Model & Architecture Type & Pretraining Used & Inria  & DG & Composite  & Parameters (M) \\  
            \midrule 
            \midrule 
            \rowcolor[gray]{0.95}
            TransUnet  &    Fully Connected    & No  & 76.07 & 77.75 &  76.99 & 103.76 \\
            TransUnet  &    Convolutional      & No  & 77.42 & 79.48 &  78.55 & 105.82 \\
            \rowcolor[gray]{0.95}
            TransUnet  &    Transformer        & No  & 77.88 & 79.66 &  78.86 & 105.91 \\
            TransUnet  &    Transformer        & Yes & 79.22 & 80.56 &  79.96 & 105.91 \\
            \bottomrule 
        \end{tabular} 
    }} 
\end{center} 
\vspace{-0.5em}
\caption{We explore the effect of using transformer layers in the last stage of the encoder. Here we compare the standard TransUnet model with variations that replace the transformer layers with fully connected layers or $3\time 3$ convolutional blocks. Note that pretrained weights are used for all layers except the transformer layers and their replacements.} 
\label{TranResults}
\vspace{-1.5em}
\end{table*} 
\setlength{\tabcolsep}{6pt}

\setlength{\tabcolsep}{5pt}
\begin{table} [tp]
  \begin{center} 
    {\small{ 
        \begin{tabular}{|l|c|c|c|c|c|} 
            \toprule 
            \multirow{2}{*}{Model} & Hyperparameters  & \multirow{2}{*}{Inria}  & \multirow{2}{*}{DG} & \multirow{2}{*}{Composite}   \\ 
                                   &  Used  &   &      &     \\ 
            \midrule 
            \midrule 
            \rowcolor[gray]{0.95}
            \multirow{2}{*}{TransUnet}  &    TransUnet   & 79.22 & 80.56 &  79.96  \\ 
              &    Unet        & 79.65 & 80.91 &  80.35 \\
            \hline
            \rowcolor[gray]{0.95}
            \multirow{2}{*}{Unet}  &    TransUnet        & 78.57 & 80.36 &  79.55 \\
              &    Unet             & 78.41 & 80.46 &  79.53 \\ 
            \bottomrule 
        \end{tabular} 
    }} 
\end{center} 
\vspace{-0.5em}
\caption{We tested the effects of using hyperparameters found from a BO search on one model  when applied to another model. For example we used the hyperparameters found through our BO procedure on Unet and trained a TransUnet with those parameters. The table shows that performance is somewhat robust to the hyperparameters used for training. The "Hyperparameters Used" column denotes which model was used in the hyperparameter search.} 
\label{BOResults}
\vspace{-0.5em}
\end{table} 
\setlength{\tabcolsep}{6pt}

\subsection{Layer Ablation}
\label{sec:layer_ablation}
We considered the effect that the number of transformer layers used had on performance. Table \ref{LayerResults} shows the performance of the TransUnet with $0$, $6$, $8$, $10$, and $12$ transformer layers. The Unet is included for comparison. We found that on average the TransUnet with $8$ transformer layers performed the best but that the performance did not change dramatically for any model other than the $0$ layer TransUnet. We also found that the $12$ layer TransUnet performed the worst of all of the TransUnet models with transformer layers. This leads us to believe that the BO search found generally good hyperparameters that work well across different variations of the TransUnet. 

Another interesting finding was that the TransUnet with $0$ transformer layers performed almost as well as it did with transformer layers; it outperformed the RestNet34-based U-Net despite it have somewhat more parameters, and it performed equivalent to the Unet baseline model despite it having substantially more parameters. This implies that the transformer portion of the TransUnet is only one factor in its performance advantage over the Unet architecture, and that there are other factors contributing to the TransUNet's performance advantages. One notable difference between the Unet and the TransUnet is that the Unet does not use padding in many of its convolutional layers, leading to smaller feature tensors and resulting in a difference in the dimensions between the encoder and decoder. This difference in dimensions does not allow for all of the encoder features to be passed to the decoder via the skip connections and reduces the amount of high resolution information provided to the decoder. While we did not test this hypothesis, which would require large architectural changes to either the TransUnet or the Unet, there were few other factors that we believe could cause such a large difference in performance.

\subsection{Pretraining and Parameter Effect}
It is well known that pretraining is important for good model performance, this is especially important for transformer models \cite{dosovitskiy2020image}. We evaluated the impact that pretraining has by training a model with randomized weights for the transformer layers only. Table \ref{TranResults} displays the performance of the TransUnet with and without pretraining for its transformer layer. Note that pretrained weights were still used for all other layers of the model. The randomized TransUnet underperforms even the TransUnet with no transformer layer, indicating that randomizing the weights is more detrimental to the performance than not including the layers.

We also investigated the importance of using BO for each model architecture was. It is very time consuming to optimize for every architectural change in the model and cross architecture comparisons become moot if there is large variability in the performance depending on how the hyper parameters are obtained. To test this we use the hyperparameters found from the Unet BO search to train the TransUnet and vice versa. Table \ref{BOResults} shows that performance is consistent in this scenario and actually improves for both models.



\begin{figure}[tp] 
\begin{center} 
\fbox{ \includegraphics[width=0.70\columnwidth]{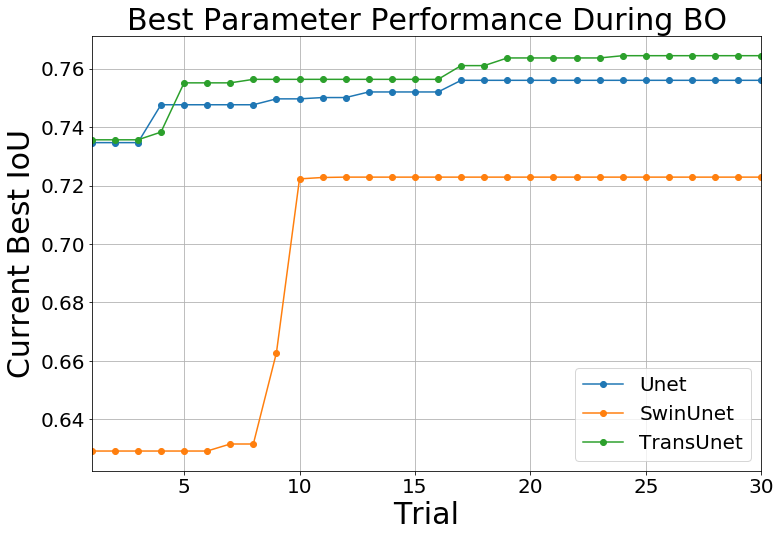} }
\vspace{-0.5em}
\end{center} 
   \caption{The max IoU trial versus the current iteration is displayed for all three baseline Bayesian optimizations. Each model improves greatly in the beginning but then converges at around or before the 20th iteration, with minor or no improvements in performance afterwards. } 
\label{fig:BO_RunningMaxIoU} 
\vspace{-.5em}
\end{figure}

\begin{figure}[tp] 
\begin{center} 
\fbox{ \includegraphics[width=0.27\columnwidth]{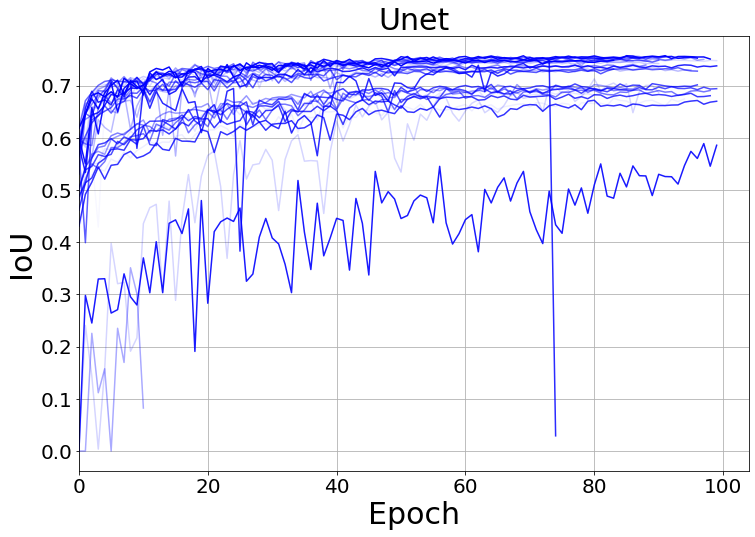} }
\fbox{ \includegraphics[width=0.27\columnwidth]{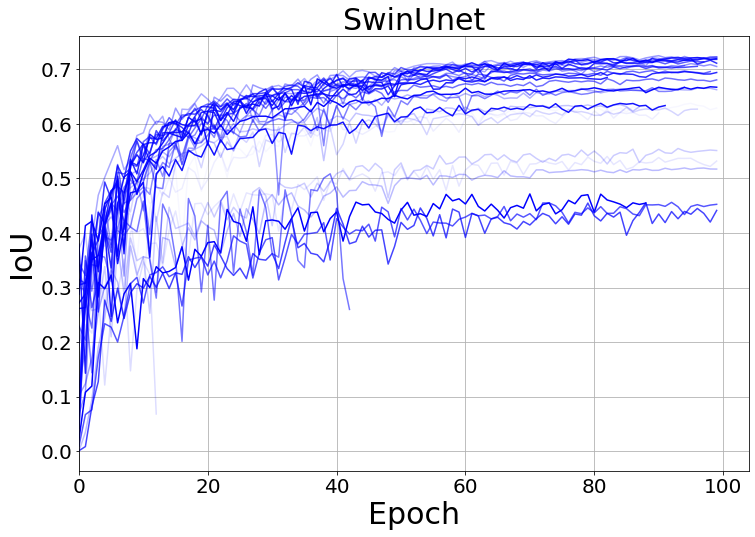} }
\fbox{ \includegraphics[width=0.27\columnwidth]{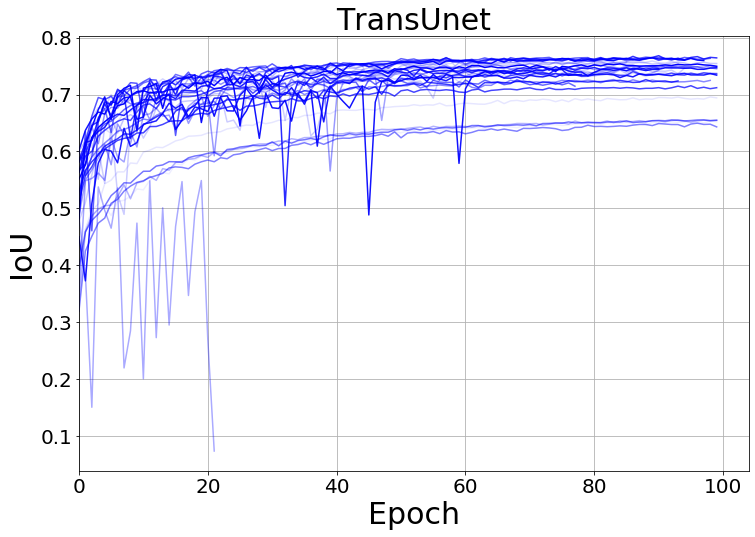} }
\vspace{-0.5em}
\end{center} 
   \caption{Shown here is the training validation at every epoch for each iteration of Bayesian optimization. Each model was allowed $30$ iteration of Bayesian optimization to find the optimum hyperparameters. The darker lines represent the later trials and the lighter the line, the earlier the trial. Since we used an adaptive stopping criteria some trials end earlier if their performance stagnates or is very poor for that stage of training. } 
\label{fig:BO_AllTraining} 
\vspace{-1.5em}
\end{figure}


\subsection{Transformer Effect}
\label{subsec:transformer_effect}
In discovering that the transformer layer is not necessary for improvement over the baseline Unet, we explored some variations of the baseline TransUnet model by replacing the transformer layer with other layer types. Since these modifications are non-standard we did not have pretrained weights for these layers and thus randomly initialized them. To account for this, we compare the modified models to the TransUnet with pretrained and randomized weights. This is not a perfect comparison since transformer models seem to be more sensitive to pretraining in general. 

First, we simply replaced the transformer layers with fully connected layers. The input to these fully connected layers were the same patch embedding used for the transformer layers. The authors of Unet motivate using transformer layers to help gather global context in the encoding process. Fully connected layers should be able to also model global relationships in the image. The added fully connected layers are of the same input and output dimensions as the input embedding. We used ReLU activation \cite{nair2010rectified} and layer normalization \cite{ba2016layer} between every fully connected layer. Enough layers were added such that the total parameter count was similar to our standard TransUnet model.

We also evaluated the effect of replacing the transformer layers with more convolutional layers. We removed the linear embedding as to retain the spatial dimensions of the features. We did not explore many different configurations for the convolutional replacements and just used simple $3 \times 3$ convolutional filters with the same channel size as the transformer layers and residual connections between layers. We used ReLU activation and batch normalization \cite{ioffe2015batch} after each convolutional layer. To keep the comparisons as fair as possible we added enough layers to increase the parameter count to be equal to the standard TransUnet.

The results presented in Table \ref{TranResults} show that using transformer layers has a greater impact than naively using other layer types. While it is impossible to say how much larger of an impact transformers have on performance compared to using convolutional or fully connected layers without an exhaustive search over the architectural space, we have found that using transformers provided slight performance improvements over using convolutions or fully connected layers in a reasonable manner.

\subsection{Effectiveness of Bayesian Optimization}
\label{subsec:effectiveness_of_bo}
The conclusions of this work depends heavily upon the premise that BO found good hyperparameters for each of our competing models, reflecting their performance in practice, given a typical systematic optimization of model hyperparameters. In this section we provide evidence that \textit{three} key steps of our BO was effective, providing strong evidence the BO process as a whole were effective.  

First we present evidence that BO found near-optimal parameters within its search range.  To do this, for each of our three BOs (one for each competing model in our experiments), we report the model's performance (IOU) estimated by the Gaussian Process model as a function of hyperparameter settings.  These estimates are made over a dense grid and are reported as an image in Fig. \ref{fig:BO_Heatmap}, where we have also overlaid the IoUs obtained by experiment at hyperparameter settings sampled by the BO. From these results, we see that there was a clear local optimum for each of the three models, suggesting that we chose sufficiently large search ranges for each model to find good hyperparameters. Furthermore, the BO sampled one (or more) points near to these local optima, suggesting that near-locally-optimal hyperparameters were obtained for each model. 

These conclusions are corroborated by results presented in Fig. \ref{fig:BO_RunningMaxIoU}, where we also report the maximum IoU obtained as a function of the number of BO iterations run for each model. We see that for each model the IoUs initially found were relatively low, and then (often steadily) increased until reaching some point of saturation, where greater IoUs were not found after many iterations. It is clear from Fig \ref{fig:BO_Heatmap} too that the BO models were not simply sampling similar settings (which is possible in BO with poor BO hyperparameter settings), but instead they sampled a diverse set of hyperparameters across the search space.  These results suggest that BO effectively improved the hyperparameter settings, and did so until a robust optimum setting was found.     

Collectively, these results suggest that BO was effective, as long as the IoUs obtained from individual experiments (i.e., training and validating a model with a single hyperparameter) were valid. This is not guaranteed, since we needed to carefully design an automatic stopping criteria for training the models, which allowed models to train until they consistently did not improve, and before any overfitting reduces their performance.  In Fig. \ref{fig:BO_AllTraining} we report the validation IOU as a function of epoch for all of the models trained during the BO process, where most models exhibit expected validation error during training, and appear to saturate in IOU at, or before, the end of training.

\lvv
\section{Conclusion}
\label{sec:conslusions}
In this work we studied whether, and to what degree, transformers are beneficial for segmentation tasks in overhead imagery.  To address this question, we performed a large-scale systematic empirical comparison of three state-of-the-art segmentation models, where each model utilizes progressively more transformer-based structures.  We considered the following three models: U-Net \cite{Iglovikov_2018_CVPR_Workshops}, TransUnet \cite{chen2021transunet}, and SwinUnet \cite{cao2021swin}. Based upon our results, we make the following conclusions:
\begin{itemize}
    \item\lv Transformers provide consistent, but modest, performance improvements.  This performance advantage was only observed in hybrid architectures (e.g., the TransUnet), comprising a convolutional encoder followed by transformers, which performed best among all models.  Fully transformer-based models (e.g., SwinUnet) achieved relatively poor performance. 
    \item\lv We found that the U-Net structure used in the Trans-U-Net (e.g., when operated without any transformer layers) performed better than other U-Net structures of comparable size (see Section \ref{sec:layer_ablation}). We were unable to isolate the precise cause of this advantage. \lv
\end{itemize}
To our knowledge, this represents the most systematic comparison of transformers within the remote sensing literature to date, providing more robust evidence regarding the impact of transformers. We note several limitations of our experiments that should be considered when interpreting our results: (i) we focused on segmentation tasks; (ii) we only employed building target classes; (iii) we used relatively small training sets compared to color imagery applications.  




\lvv
\section*{Acknowledgment}
We thank the Energy Initiative at Duke University for their support.  This work was supported in part by the Alfred P. Sloan Foundation. The content is solely the responsibility of the authors and does not necessarily represent the official views of the Alfred P. Sloan Foundation.

\clearpage
\linespread{1.0}
{\small 
\bibliographystyle{ieee_fullname} 
\bibliography{Main} 
}

\end{document}